\documentclass[11pt]{article}
\usepackage{times}
\usepackage{latexsym}
\usepackage[T1]{fontenc}
\usepackage[utf8]{inputenc}
\usepackage{microtype}
\usepackage{inconsolata}
\usepackage{graphicx}
\usepackage{xcolor}
\usepackage{amsmath}
\usepackage{booktabs}
\usepackage{array}
\usepackage{url}
\usepackage{fancyvrb}
\usepackage{fontawesome5}
\usepackage{hyperref}
 \usepackage[final]{acl}  % camera-ready
\hypersetup{
    colorlinks=true,
    urlcolor=blue!80!black
}
\newcolumntype{L}[1]{>{\raggedright\arraybackslash}p{#1}}
\newcommand{\muse}{\textsc{MUSE}}
\newcommand{\psr}{\mbox{P--S--R}}
\providecommand{\ps}{}\renewcommand{\ps}{\texttt{problem\_salient}}
\providecommand{\ss}{}\renewcommand{\ss}{\texttt{solution\_salient}}
\title{\muse: A Full-Text Cross-Domain Knowledge Base of Scientific Problems, Solutions, and Rationales}

\author{
   Tsofia Cohen$^1$ \quad Tom Hope$^{1,2}$ \\[0.1cm]
    $^1$The Hebrew University of Jerusalem \\
    $^2$Allen Institute for AI (Ai2) \\[0.1cm]
    \texttt{\{tsofia.cohen, tom.hope\}@mail.huji.ac.il} \\[0.2cm]
    \href{https://github.com/cohentsofia/MUSE}{\faGithub~GitHub} \quad
    \href{https://huggingface.co/datasets/TsofiaCohen/MUSE}{\faDatabase~Data} \quad
    \href{https://huggingface.co/collections/TsofiaCohen/muse}{\faCubes~Models}
}

\date{}

\begin{document}
\maketitle

\begin{abstract}
Scientific papers contain fine-grained records of problem solving: authors describe technical obstacles, the methods used to address them, and often their \textit{reasoning} about why those methods were chosen. We introduce \muse{} (Mining Underlying Scientific Explanations), a full-text, multi-domain resource of scientific Problem--Solution--Rationale (\psr{}) triplets. We curate expert-annotated full-text paragraphs with a rich schema of salient spans, problem--solution and rationale relations, and conceptual coreference, and train a modular extraction pipeline that scales this annotation to a source-grounded knowledge base of 32{,}557 triplets across arXiv. We evaluate each pipeline component, audit the resulting knowledge base end-to-end, and release it publicly. Finally, a preliminary study shows that \psr{} triplets can serve as a training signal for scientific problem solving: supervising on rationales improves solution quality over solution-only supervision for one of the two models we study.
\end{abstract}

\section{Introduction}
\label{sec:intro}

\begin{figure*}[t]
    \centering
\includegraphics[width=1.0\linewidth]{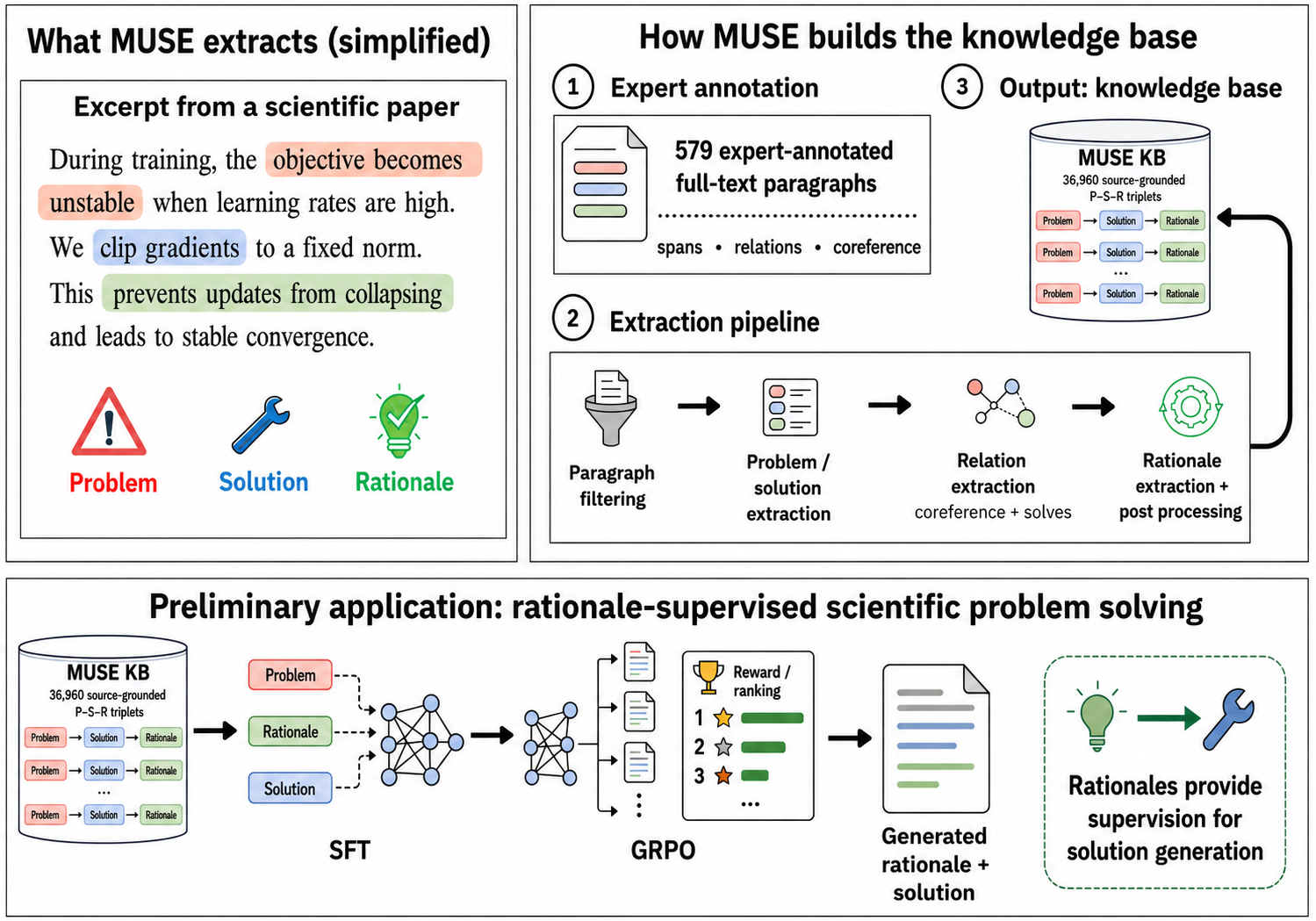}
    \caption{The \muse{} extraction pipeline and one application. \textbf{Top left:} the three core entity types. \textbf{Top right:} the construction pipeline, resulting in 32,557 \psr{} triplets. \textbf{Bottom:} a preliminary downstream application, where rationale-supervised fine-tuning tests whether the rationale field provides a useful training signal beyond solution-only supervision via SFT followed by GRPO.}
    \label{fig:PRS_OVERVIEW}
\end{figure*}

A fundamental goal in Scientific NLP is to convert scholarly text into structured representations that support retrieval, synthesis, and computational analysis. This objective has motivated construction of large scholarly corpora \citep{lo2020s2orc,unarxive2023}, as well as information-extraction (IE) benchmarks for scientific keyphrases, typed entities, semantic relations, coreference, and document-level n-ary relations \citep{augenstein2017scienceie,scierc2018,scirex2020}.

Full-text resources annotate entities and relations in complete papers, such as dataset, method, task and metric \citep{scier2024,scinlp2025}, and discourse-oriented resources annotate rhetorical or communicative roles such as background, motivation, method, result, and conclusion \citep{liakata2010coresc,scidtb2018,huang2020coda}. Scientific knowledge bases (KBs) built on IE include problems and solutions, contributions, challenges, and inspirations \citep{heffernan2018problems,jansen2026scg,chimera2025,lahav2022}.

These resources support many important tasks, but they do not directly represent a common structure in scientific papers: a fine-grained concrete technical sub-problem (which can appear deep in the full text, not being the main problem addressed by the paper), the solution used for that problem, and the \emph{rationale} that explains why the solution is appropriate. This is more specific than a paper-level contribution---indeed, the problems and solutions may not even be part of the paper's contribution per se.

We introduce \muse{} (Mining Underlying Scientific Explanations), a dataset and knowledge base for full-text scientific Problem--Solution--Rationale (\psr{}) extraction. In \muse{}, a \emph{problem} is a limitation, obstacle, gap, or desideratum; a \emph{solution} is a method, mechanism, formulation, design choice, or procedure that was used to address it; and a \emph{rationale} is the author-stated motivation, justification or explanation connecting the solution to the problem. The annotation scheme captures both textual evidence and relational structure: sentences and salient spans for problems, solutions, and rationales; \texttt{solves} links; \texttt{rationale\_of} links; and conceptual coreference among fragmented or repeated mentions.

The resource has two intended contributions as an IE artifact. First, it is full-text and broad-domain: the extracted triplets are mined from paragraphs throughout full-text scientific papers from multiple domains, rather than only abstracts; this allows the resource to capture technical justifications that often appear in methods, analysis, appendix, and limitation sections. Second, it is structurally rich. Each example retains source provenance and multiple annotation layers: paragraph relevance, span boundaries, typed relations, concept-level coreference groups, and post-processed standalone problem, solution, and rationale fields. This makes \muse{} useful both as a structured KB and as training or evaluation data for IE.

Figure~\ref{fig:PRS_OVERVIEW} gives a high-level overview of the resource, the pipeline, and one downstream application. We train a dedicated extraction pipeline on \muse{} and use it to construct a KB of 32{,}557 \psr{} triplets across all domains in arXiv (categories in Table~\ref{tab:arxiv_categories}). We envisage this KB may support many applications, including training and evaluation sets for AI scientific problem solving, cross-domain ideation through analogous rationales, tools for surfacing expert reasoning patterns, and meta-scientific analyses of how fields formulate constraints and justify methods. As one demonstration, we study whether extracted rationales can supervise LLMs for scientific problem solving. For Qwen3-32B, training on the rationale together with the solution significantly improves solution quality over solution-only training, while DeepSeek-R1-Distill-Qwen-32B shows no significant change and neither fine-tuned model surpasses its base model. Where the gain occurs it is concentrated in novelty and technical depth. These experiments are preliminary and primarily illustrate an application of the resource.

Our contributions are:
\begin{enumerate}
  \item \textbf{A full-text scientific IE task and expert-annotated seed dataset} for Problem--Solution--Rationale extraction, including salient spans, \texttt{solves} links, \texttt{rationale\_of} links, and conceptual coreference.
  \item \textbf{A large structured knowledge base} of 32,557 source-grounded \psr{} triplets mined from full-text scientific papers across domains and represented as self-contained problem, solution, and rationale fields, with an end-to-end audit of uniformly sampled records and a characterization of domain coverage.
  \item \textbf{A preliminary application study} showing that, for Qwen3-32B, \muse{} rationales provide a significant training signal for scientific problem solving beyond solution-only supervision, together with a held-out moderator analysis showing that the benefit is broad rather than tied to problem complexity or domain, and does not come from over-engineered solutions.
\end{enumerate}

\begin{figure*}
    \centering
\includegraphics[width=1.0\linewidth]{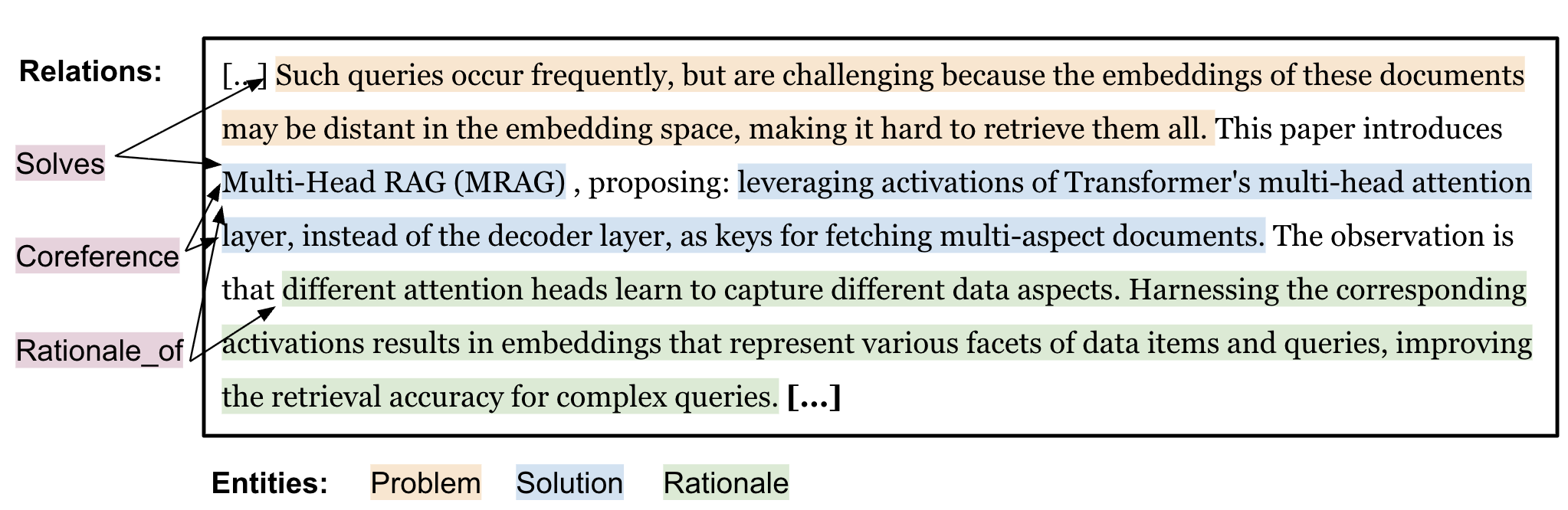}
    \caption{The \muse{} schema extracts scientific rationale--problem-solving structure from full-text paragraphs, including salient spans and conceptual coreference relations.}
    \label{fig:annotation_example_fig}
\end{figure*}

\section{Related Work}
\label{sec:related}

\paragraph{Scientific information extraction.}
Scientific IE involves abstract-level or paper-level entity and relation extraction tasks. ScienceIE focuses on scientific keyphrases and semantic relations such as synonym and hyponym links \citep{augenstein2017scienceie}. SciERC annotates scientific entities, relations, and coreference clusters in abstracts \citep{scierc2018}. SciREX targets document-level IE over scientific articles, including salient entity identification and n-ary relation extraction \citep{scirex2020}. Recent full-text resources such as SciER and SciNLP extend this line to complete papers, annotating entities and relations involving datasets, methods, tasks, metrics, and other scientific artifacts \citep{scier2024,scinlp2025}.

Discourse-oriented resources represent the rhetorical function of scientific passages. CoreSC-style annotation identifies conceptual zones such as background, motivation, goal, method, result, and conclusion \citep{liakata2010coresc}; SciDTB annotates discourse dependencies in scientific abstracts \citep{scidtb2018}; and CODA-19 scales research-aspect annotation to COVID-19 abstracts \citep{huang2020coda}.

\citet{heffernan2018problems} create a corpus of positive and negative problem and solution phrases and train classifiers for problemhood and solutionhood. \citet{lahav2022} annotate challenges and research directions in full-length COVID-19-related papers and build a search engine for scientific discovery.

\muse{} differs in several ways. \muse{} extracts problem-solving structures and the rationale connecting solutions to the problems they address, in full-text passages, from many domains. It operates at paragraph and span level rather than phrase or sentence level; and it links solutions to the specific problems they address and adds rationales and rationale relations.

\paragraph{Scientific knowledge bases and discovery systems.}
Scientific KB construction has increasingly moved toward richer representations of contributions. CHIMERA mines idea recombinations and inspirations for scientific ideation \citep{chimera2025}; the Scientific Contribution Graph models contributions and prerequisite relations for literature-based technological roadmapping \citep{jansen2026scg}. \muse{} is complementary: it extracts local problem--solution--rationale units that can be indexed, compared, and reused. This reuse motivation also connects to recent work on adaptive and agentic reasoning: \citet{ranaldi2026evolving} introduces semi-structured, reusable abstractions of problem-solving mechanisms that help LLM agents adapt and evolve their reasoning over time, and \muse{}'s extracted \psr{} units offer exactly this kind of explicit, literature-grounded representation of how a solution addresses a problem and why.

\paragraph{Rationales and explanations in NLP.}
Natural-language rationales have been studied in interpretability and explanation datasets, including e-SNLI for natural language inference explanations \citep{camburu2018esnli} and ERASER for evaluating rationalized NLP models across tasks \citep{deyoung2020eraser}. In contrast, \muse{} rationales are not annotations of why a model predicted a label; they are author-written scientific justifications that explain why a technical solution addresses a technical problem. This makes them useful for scientific search, problem-solving evaluation, and analysis of expert justification patterns, and potentially as additional supervision for training reasoning LLMs.

\section{The \psr{} Task and Dataset}
\label{sec:task}

We first formalize the Problem--Solution--Rationale extraction task and then describe the expert-annotated seed dataset used to train and evaluate the extraction pipeline. Figure~\ref{fig:annotation_example_fig} illustrates the annotation schema applied to a representative full-text paragraph.

\subsection{Task Definition}
Given a scientific paragraph $D$, the task is to recover a set of \psr{} structures. Each structure contains:
\begin{itemize}
  \item \textbf{Problem}: a concrete limitation, obstacle, gap, or desideratum;
  \item \textbf{Solution}: a method, mechanism, design choice, formulation, or procedure used to address the problem;
  \item \textbf{Rationale}: the stated motivation, justification or explanation why the solution was chosen to address the problem.
\end{itemize}

At the annotation level, problems and solutions are marked as salient spans, \ps{} and \ss{}, and rationales are marked as \texttt{rationale} spans. We additionally annotate three relations:
\texttt{solves} links a solution to the problem it addresses;
\texttt{rationale\_of} links a rationale to the solution it explains; and
\texttt{coreference} links multiple spans that refer to the same underlying problem or solution concept. Table~\ref{tab:layers} summarizes the layers captured by the annotation and post-processing scheme. A detailed annotation codebook---with the operational definitions, inclusion/exclusion criteria, and boundary rules used by annotators---is provided in Appendix~\ref{app:annotation}.

\begin{table}[t]
  \centering
  \small
  \setlength{\tabcolsep}{3pt}
  \begin{tabular}{p{0.28\linewidth}p{0.62\linewidth}}
    \toprule
    Layer & Captured information \\
    \midrule
    Paragraph relevance & Whether a paragraph contains problem--solution structure and rationale evidence \\
    Salient spans & Minimal problem, solution, and rationale spans grounded in source text \\
    Typed relations & \texttt{solves} and \texttt{rationale\_of} links among spans \\
    Coreference & Conceptual groups of fragmented or repeated problem/solution mentions \\
    \bottomrule
  \end{tabular}
  \caption{Structural layers represented in \muse{}.}
  \label{tab:layers}
\end{table}

\subsection{Annotated Seed Dataset}
\label{sec:seed}
We sample full-text paragraphs from arXiv (via unarXive; see Appendix~\ref{app:annotation}), using high-recall keyword filters to make annotation more efficient than random sampling of full-text paragraphs. The seed set contains 579 manually annotated paragraphs, comparable to or larger than sets used in related scientific IE resources (e.g., CHIMERA \citep{chimera2025} and CARE \citep{naik2024care}). Unlike these abstract-level resources, we annotate full-text paragraphs. Annotation was conducted by a PhD-level scientific annotator after screening and training; a second NLP expert independently annotated a held-out quality-control subset, disagreements were discussed, guidelines were refined, and the annotator revised examples accordingly. Agreement was high (first row of Tables~\ref{tab:psr-cls}--\ref{tab:rel}). We use an 80\%/10\%/10\% train/validation/test split at the paper level. We note as a limitation (\S\ref{sec:limitations}) that annotation relies on a single primary annotator with expert review rather than multiple independent annotators; additional details on sourcing, the codebook, and metrics appear in Appendix~\ref{app:annotation}.

\section{The \muse{} Extraction Pipeline}
\label{sec:pipeline}

\begin{figure*}
    \centering
\includegraphics[width=1.0\linewidth]{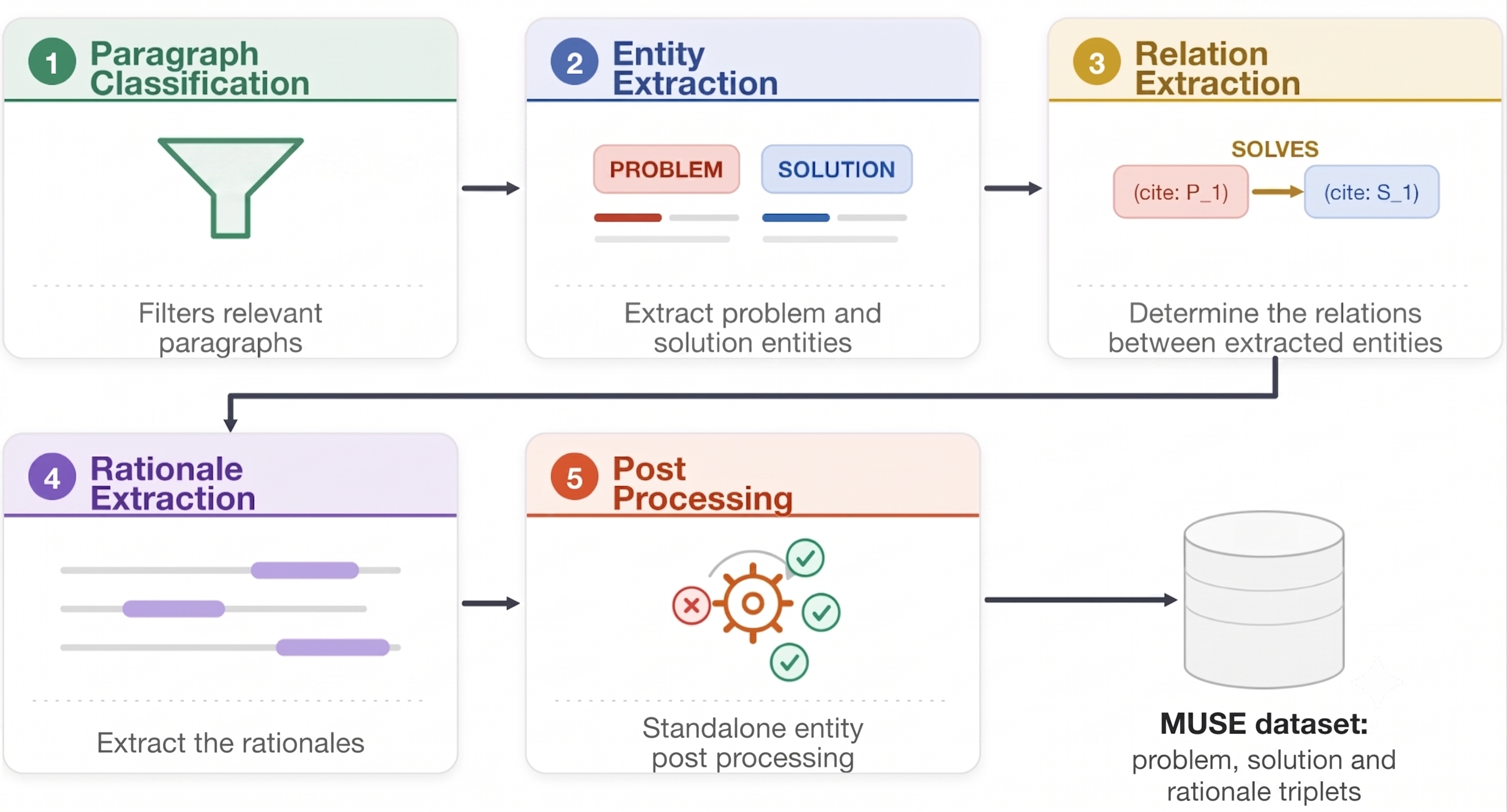}
    \caption{High-level overview of the \muse{} extraction pipeline.}
    \label{fig:muse_pipeline_fig}
\end{figure*}

Figure~\ref{fig:muse_pipeline_fig} gives a high-level overview. The design differentiates between labels that are local and boundary-based, and outputs that require conceptual grouping or source-grounded rewriting.

\paragraph{Why not end-to-end LLM extraction?} Before designing the modular pipeline, we explored prompting a single LLM to extract complete \psr{} structures from a paragraph in one pass. On simple paragraphs with a single problem and solution the model was often adequate, but it degraded sharply on the structured cases that motivate this work: it frequently conflated solutions with their rationales, dropped the relations between multiple problem--solution pairs, and failed to preserve the correspondence among problems, solutions, and rationales when a paragraph interleaved several concepts. Rationale extraction was particularly brittle---given only the paragraph, the model tended to summarize the paper's overall contribution rather than the local justification for a specific solution (Appendix~\ref{app:end2end}). These failures motivate our decomposition into paragraph filtering, span extraction, relation extraction, and post-processing, so that each stage solves a narrower, better-posed problem; in particular, rationale construction is conditioned on already-extracted problem and solution anchors rather than on the raw paragraph alone.

\paragraph{Paragraph filtering.}
The first stage identifies paragraphs likely to contain a problem, a solution, and a rationale. A single holistic classifier was unreliable on the small seed set, so we use a cascade: first detect Problem--Solution paragraphs, then detect whether a rationale is present among those paragraphs. For large-scale construction, each sample receives a probability score from the cascade; samples scoring above $0.5$ are classified as relevant. To prioritize precision during KB construction, we retain only the top 10\% of paragraphs ranked by the cascaded classifier's confidence score.

\paragraph{Problem and solution extraction.}
For paragraphs that pass filtering, we extract \ps{} and \ss{} spans with token-level classifiers. We evaluate multiple encoder architectures and labeling formulations, and select the best-performing multi-class DeBERTa-v3-large model \citep{deberta2021} for both problem and solution spans. Rationale spans are substantially less boundary-stable: they are often distributed across clauses or require synthesis. We therefore do not rely on token classification for the final rationale field.

\paragraph{Relation extraction.}
We fine-tune Mistral-7B-Instruct \citep{mistral2023} for two relational steps. First, conceptual coreference groups problem or solution spans that refer to the same underlying concept. Second, a \texttt{solves} classifier links each solution group to the problem group it addresses. This produces structured problem--solution candidates with provenance in the original paragraph.

\paragraph{Post-processing and rationale construction.}
The final fields in the KB must be readable outside the source paragraph. We prompt an LLM (GPT-4o) with the paragraph and extracted spans to produce concise, self-contained problem and solution descriptions. We then prompt GPT-4o with the paragraph and the post-processed problem--solution pair to extract a faithful, source-grounded rationale. This step rewrites and contextualizes the author-described rationale rather than treating the generated text as an independent model explanation. The span-level output is post-processed into a self-contained triplet
\[
t = (\text{problem}: P,\ \text{solution}: S,\ \text{rationale}: R),
\]
where $P$, $S$, and $R$ can be read independently of the source paragraph while preserving provenance.

\section{Extraction Results}
\label{sec:extraction-results}

We evaluate each component of the pipeline against the expert annotations, proceeding from paragraph filtering through span extraction to relation extraction and post-processing. In every results table, the top row reports human agreement as an approximate upper bound.

\subsection{Paragraph Classification}
The holistic \psr{} classifier performs poorly on the seed set (Table~\ref{tab:psr-cls}), motivating the cascaded design. The cascaded classifier (Table~\ref{tab:cascade}) improves the relevant-class $F_1$, and the top-confidence version prioritizes precision for KB construction.

\begin{table}[h]
  \centering
  \small
  \setlength{\tabcolsep}{4pt}
  \begin{tabular}{lccc ccc}
    \toprule
    & \multicolumn{3}{c}{relevant} & \multicolumn{3}{c}{irrelevant} \\
    \cmidrule(lr){2-4}\cmidrule(lr){5-7}
    Model & P & R & F1 & P & R & F1 \\
    \midrule
    Human-Agreement & 0.89 & 0.94 & 0.91 & 0.98 & 0.95 & 0.96 \\
    \midrule
    SciDeBERTa & 0.20 & 0.19 & 0.19 & 0.70 & 0.71 & 0.71 \\
    DeBERTa-v3-base & 1.00 & 0.06 & 0.12 & 0.74 & 1.00 & 0.85 \\
    RoBERTa-base & 0.00 & 0.00 & 0.00 & 0.72 & 1.00 & 0.84 \\
    Cs-RoBERTa & 0.32 & 0.38 & 0.34 & 0.74 & 0.69 & 0.72 \\
    SciBERT & 0.33 & 0.19 & 0.24 & 0.74 & 0.86 & 0.79 \\
    DistilBERT & 0.15 & 0.25 & 0.19 & 0.61 & 0.45 & 0.52 \\
    Mistral-7B v0.3 & 0.39 & 0.31 & 0.35 & 0.78 & 0.76 & 0.77 \\
    \bottomrule
  \end{tabular}
  \caption{Holistic \psr{} paragraph classification.}
  \label{tab:psr-cls}
\end{table}

\begin{table}[h]
  \centering
  \small
  \setlength{\tabcolsep}{4pt}
  \begin{tabular}{lccc ccc}
    \toprule
    & \multicolumn{3}{c}{relevant} & \multicolumn{3}{c}{irrelevant} \\
    \cmidrule(lr){2-4}\cmidrule(lr){5-7}
    Classifier & P & R & F1 & P & R & F1 \\
    \midrule
    Human-Agreement & 0.89 & 0.94 & 0.91 & 0.98 & 0.95 & 0.96 \\
    \midrule
    Cascaded & 0.74 & 0.71 & 0.73 & 0.72 & 0.75 & 0.74 \\
    Cascaded @10 & 1.00 & 0.58 & 0.73 & 0.70 & 1.00 & 0.82 \\
    \bottomrule
  \end{tabular}
  \caption{Cascaded paragraph classification with Mistral-7B components.}
  \label{tab:cascade}
\end{table}

\subsection{Entity Extraction}
Encoder classifiers are evaluated with strict token-level matching; candidates include DeBERTa-v3 \citep{deberta2021}, RoBERTa \citep{roberta2019}, SciBERT \citep{scibert2019}, and DistilBERT \citep{distilbert2019}. The generative Mistral extractor is evaluated with thresholded SciBERT BERTScore. Table~\ref{tab:entity} shows that multi-class DeBERTa-v3-large is strongest for problem and solution spans, while rationale spans are not reliably captured by token classification---motivating the LLM-based, anchor-conditioned rationale construction used for the final field.

\begin{table*}[t]
  \centering
  \footnotesize
  \setlength{\tabcolsep}{4.5pt}
  \begin{tabular}{lccc ccc ccc}
    \toprule
    & \multicolumn{3}{c}{\ps{}} & \multicolumn{3}{c}{\ss{}} & \multicolumn{3}{c}{\texttt{rationale}} \\
    \cmidrule(lr){2-4}\cmidrule(lr){5-7}\cmidrule(lr){8-10}
    Model & P & R & F1 & P & R & F1 & P & R & F1 \\
    \midrule
    Human-Agreement & 0.85 & 0.85 & 0.85 & 0.80 & 0.93 & 0.86 & 0.70 & 0.72 & 0.71 \\
    \midrule
    \multicolumn{10}{l}{\emph{Multi-Label Token Classification}} \\
    SciDeBERTa       & 0.35 & 0.79 & 0.48 & 0.35 & 0.85 & 0.50 & 0.12 & 0.33 & 0.18 \\
    DeBERTa-v3-base  & 0.63 & 0.65 & 0.64 & 0.57 & 0.61 & 0.59 & 0.19 & 0.17 & 0.19 \\
    DeBERTa-v3-large & 0.57 & 0.65 & 0.61 & 0.54 & 0.67 & 0.60 & 0.22 & 0.39 & 0.28 \\
    RoBERTa-base     & 0.43 & 0.56 & 0.49 & 0.37 & 0.69 & 0.48 & 0.08 & 0.35 & 0.13 \\
    Cs-RoBERTa       & 0.54 & 0.58 & 0.56 & 0.58 & 0.62 & 0.60 & 0.22 & 0.22 & 0.22 \\
    SciBERT          & 0.51 & 0.56 & 0.54 & 0.64 & 0.62 & 0.63 & 0.08 & 0.11 & 0.09 \\
    DistilBERT       & 0.45 & 0.51 & 0.48 & 0.44 & 0.61 & 0.51 & 0.04 & 0.13 & 0.07 \\
    \midrule
    \multicolumn{10}{l}{\emph{Multi-Class Token Classification}} \\
    SciDeBERTa       & 0.66 & 0.48 & 0.56 & 0.65 & 0.57 & 0.61 & 0.14 & 0.06 & 0.09 \\
    DeBERTa-v3-base  & 0.63 & 0.39 & 0.48 & 0.58 & 0.57 & 0.58 & 0.00 & 0.00 & 0.00 \\
    \textbf{DeBERTa-v3-large} & \textbf{0.75} & \textbf{0.58} & \textbf{0.66} & \textbf{0.67} & \textbf{0.61} & \textbf{0.64} & 0.19 & 0.10 & 0.13 \\
    RoBERTa-base     & 0.63 & 0.46 & 0.53 & 0.57 & 0.62 & 0.60 & 0.09 & 0.01 & 0.02 \\
    Cs-RoBERTa       & 0.72 & 0.37 & 0.49 & 0.52 & 0.54 & 0.53 & 0.29 & 0.00 & 0.01 \\
    SciBERT          & 0.65 & 0.36 & 0.46 & 0.66 & 0.50 & 0.57 & 0.06 & 0.02 & 0.03 \\
    DistilBERT       & 0.69 & 0.32 & 0.43 & 0.52 & 0.44 & 0.47 & 0.00 & 0.00 & 0.00 \\
    \midrule
    \multicolumn{10}{l}{\emph{Binary Token Classification}} \\
    SciDeBERTa       & 0.74 & 0.48 & 0.59 & 0.73 & 0.45 & 0.56 & 0.09 & 0.03 & 0.04 \\
    DeBERTa-v3-base  & 0.64 & 0.37 & 0.47 & 0.60 & 0.54 & 0.57 & 0.09 & 0.03 & 0.04 \\
    DeBERTa-v3-large & 0.72 & 0.50 & 0.59 & 0.67 & 0.59 & 0.63 & 0.04 & 0.01 & 0.02 \\
    RoBERTa-base     & 0.69 & 0.37 & 0.48 & 0.55 & 0.53 & 0.54 & 0.00 & 0.00 & 0.00 \\
    Cs-RoBERTa       & 0.68 & 0.38 & 0.49 & 0.59 & 0.52 & 0.55 & 0.02 & 0.00 & 0.00 \\
    SciBERT          & 0.62 & 0.25 & 0.36 & 0.73 & 0.46 & 0.57 & 0.07 & 0.02 & 0.03 \\
    DistilBERT       & 0.61 & 0.30 & 0.41 & 0.62 & 0.35 & 0.45 & 0.11 & 0.03 & 0.05 \\
    \midrule
    \multicolumn{10}{l}{\emph{Generative LLM NER}} \\
    Mistral-7B v0.3  & 0.61 & 0.25 & 0.36 & 0.72 & 0.29 & 0.43 & 0.00 & 0.00 & 0.00 \\
    \bottomrule
  \end{tabular}
  \caption{Token-level entity extraction for \ps{}, \ss{}, and \texttt{rationale}.}
  \label{tab:entity}
\end{table*}

\subsection{Relation Extraction and Post-Processing}
For conceptual coreference, the fine-tuned Mistral-7B model achieves strong link-based performance ($F_1=0.91$; Table~\ref{tab:rel}), indicating it can reliably group fragmented or repeated mentions of the same concept. The \texttt{solves} classifier performs well on positive links ($F_1=0.87$), though performance is lower on the negative class ($F_1=0.69$), suggesting the model is better at identifying plausible problem--solution relations than rejecting difficult non-relations.

The GPT-4o post-processing stage produces highly self-contained and faithful problem and solution descriptions (Table~\ref{tab:pp}): problems score 4.97 for standalone quality and 4.52 for fidelity, solutions 4.82 and 4.51. Completeness is lower, particularly for solutions (3.53), reflecting the challenge of preserving all technical details when condensing complex solution spans. Rationale construction is generally grounded and coherent (groundedness 4.70, coherence 4.84; SciBERT BERTScore $F_1=0.78$ against human rationales; Table~\ref{tab:rationale}), but specificity and completeness are more moderate, indicating rationale generation is the least mature component.

\begin{table}[h]
  \centering
  \small
  \begin{tabular}{llccc}
    \toprule
    Task & Class & P & R & F1 \\
    \midrule
    Coreference & link-based & 0.92 & 0.91 & 0.91 \\
    \midrule
    \texttt{solves} & Solves & 0.93 & 0.83 & 0.87 \\
    \texttt{solves} & No-Solves & 1.00 & 0.52 & 0.69 \\
    \bottomrule
  \end{tabular}
  \caption{Relation extraction with fine-tuned Mistral-7B.}
  \label{tab:rel}
\end{table}

\begin{table}[h]
  \centering
  \small
  \begin{tabular}{lccc}
    \toprule
    Type & Standalone & Fidelity & Completeness \\
    \midrule
    Problem  & 4.97 & 4.52 & 4.06 \\
    Solution & 4.82 & 4.51 & 3.53 \\
    \bottomrule
  \end{tabular}
  \caption{LLM-judge evaluation of post-processed problems and solutions (1--5).}
  \label{tab:pp}
\end{table}

\begin{table}[h]
  \centering
  \small
  \begin{tabular}{lc}
    \toprule
    \multicolumn{2}{l}{\emph{LLM-as-a-judge criteria (1--5)}} \\
    Groundedness & 4.70 \\
    Explanation Strength & 3.42 \\
    Specificity & 2.83 \\
    Coherence & 4.84 \\
    Non-Redundancy & 3.66 \\
    Completeness & 3.12 \\
    Overall Score & 3.54 \\
    \midrule
    \multicolumn{2}{l}{\emph{Similarity vs.\ human rationales}} \\
    Mean BERTScore Precision & 0.753 \\
    Mean BERTScore Recall & 0.814 \\
    Mean BERTScore F1 & 0.782 \\
    \midrule
    \multicolumn{2}{l}{\emph{Alignment with human rationales}} \\
    Mean LLM-Judge Score (1--5) & 3.38 \\
    \bottomrule
  \end{tabular}
  \caption{Evaluation of LLM-extracted rationales.}
  \label{tab:rationale}
\end{table}

\subsection{KB Construction and End-to-End Audit}
\label{sec:evaluation}

We apply the pipeline to arXiv full text across all categories (Table~\ref{tab:arxiv_categories}) through 2026, keeping the top 10\% of candidate paragraphs by cascaded-classifier confidence. After removing near-duplicate records, a final refinement step with Claude Opus~4.8 rewrites each record so that the solution describes only the approach and the rationale only its justification, since a pre-release check found the two fields often overlapped (Appendix~\ref{app:e2e-audit}). The final knowledge base contains \textbf{32,557 \psr{} triplets}, released with train/validation/test splits of 32{,}247/161/149, together with the pre-refinement fields. Each triplet is grounded in a specific source paragraph (provenance retained in the full KB) and released as standalone problem, solution, and rationale fields, so the KB is usable both as a text resource and as a structured graph. An end-to-end audit of a uniform sample against the source paragraphs finds problems and solutions reliable (91.5\% and 77.0\%). Rationales are the hardest field to recover, since authors often leave their reasoning implicit. Appendix~\ref{app:e2e-audit} reports the full audit together with KB statistics and domain coverage.

\section{Preliminary Application: Rationale-Supervised Problem Solving}
\label{sec:poc}
To illustrate one use of \muse{}, we conduct an exploratory study of rationale-supervised fine-tuning. The goal is not a state-of-the-art reasoning model, but to test whether the rationale field provides a useful training signal beyond solution-only supervision, and to characterize \emph{when} it does.

\subsection{Setup}
For two open-source models---Qwen3-32B \citep{qwen3-2025} and DeepSeek-R1-Distill-Qwen-32B \citep{deepseekr1}---we train two variants that differ only in the target:
\begin{itemize}
\item \textbf{PS:} problem $\rightarrow$ solution, omitting rationale;
\item \textbf{PRS:} problem $\rightarrow$ rationale $\rightarrow$ solution, using the full \psr{} triplet.
\end{itemize}
Both variants use the same examples and decoding settings; comparing PS to PRS therefore isolates the effect of the rationale signal. Training uses supervised fine-tuning followed by Group Relative Policy Optimization (GRPO) \citep{deepseekmath2024} with reward
\begin{equation}
\begin{split}
R_{\mathrm{total}}(x,y) = 0.7\,R_{\mathrm{corr}}(x,y)\\ + 0.1\,R_{\mathrm{fmt}}(x,y) + 0.2\,R_{\mathrm{div}}(x,y),
\label{eq:reward}
\end{split}
\end{equation}
where $R_{\mathrm{corr}}$ uses an LLM judge, $R_{\mathrm{fmt}}$ rewards parseable output, and $R_{\mathrm{div}}$ penalizes near-copying of the prompt using sentence embeddings \citep{wang2024scimon,sentencebert2019}. Hyperparameters and reward details are in Appendix~\ref{app:finetuning}. To prevent contamination, test and validation sets are drawn from arXiv papers published after the models' knowledge cutoffs; after de-duplication, 156 unique held-out problems remain, on which we report downstream results. \textbf{Evaluation} uses a blinded protocol in which Base, PS, and PRS outputs are anonymized and randomized and scored $1$--$10$ on novelty, feasibility, technical detail, overall quality, and pairwise rank by \textbf{GPT-5} as an independent LLM judge.

\subsection{Rationale supervision improves over solution-only supervision for Qwen3-32B}
\label{sec:main-effect}
Table~\ref{tab:downstream-sig} compares the two fine-tuned variants on the 156 held-out problems. For \textbf{Qwen3-32B}, rationale supervision significantly improves over the identical solution-only pipeline on all three metrics, and PRS produces the better solution on 63\% of problems versus 21\% for PS. The gain is concentrated in novelty ($+23\%$) and technical detail ($+26\%$), while feasibility changes little: learning \emph{why} a method fits makes solutions more original and substantive, not merely more plausible. For \textbf{DeepSeek-R1-Distill-Qwen-32B}, the differences are not significant, so the benefit is model-dependent.

\begin{table}[t]
\centering
\small
\setlength{\tabcolsep}{3.5pt}
\begin{tabular}{@{}lccc c@{}}
\toprule
Metric & PS & PRS & Base & $\Delta_{\text{PRS}-\text{PS}}$ (95\% CI) \\
\midrule
\multicolumn{5}{l}{\emph{Qwen3-32B}} \\
Novelty      & 3.09 & 3.80 & 4.21 & $+0.71\;[+0.42,+1.01]^{\ast\ast\ast}$ \\
Feasibility  & 6.89 & 7.18 & 7.21 & $+0.29\;[+0.07,+0.51]^{\ast\ast}$ \\
Tech.\ detail& 3.22 & 4.05 & 4.53 & $+0.83\;[+0.60,+1.06]^{\ast\ast\ast}$ \\
\midrule
\multicolumn{5}{l}{\emph{DeepSeek-R1-Distill-Qwen-32B}} \\
Novelty      & 2.65 & 2.73 & 2.65 & $+0.08\;[-0.08,+0.26]$ \\
Feasibility  & 6.34 & 6.32 & 6.34 & $-0.02\;[-0.15,+0.12]$ \\
Tech.\ detail& 2.98 & 2.96 & 3.01 & $-0.03\;[-0.19,+0.14]$ \\
\bottomrule
\end{tabular}
\caption{Downstream quality (GPT-5 judge, 1--10) on 156 held-out problems, with bootstrap 95\% CIs. PRS$-$PS significance from paired Wilcoxon tests, Holm-corrected ($^{\ast\ast}p{<}0.01$, $^{\ast\ast\ast}p{<}0.001$).}
\label{tab:downstream-sig}
\end{table}
\paragraph{The benefit is broad.} A moderator analysis with held-out replication (Appendix~\ref{app:moderators}) shows that the benefit of rationale supervision is broad, holding across problems of varying complexity and across domains.

\paragraph{Takeaway.} For Qwen3-32B, training on \muse{} rationales significantly improves scientific solution quality over solution-only training, concentrated in novelty and technical depth; for DeepSeek-R1-Distill-Qwen-32B it has no significant effect. Where the gain occurs, it is broad across problems rather than confined to any identifiable class of problems; it tracks the model's own competence and the degree of substantive elaboration rather than the intrinsic difficulty of the problem.

\section {Conclusion}
\label{sec:conclusion}
We introduced \muse{}, a full-text, cross-domain resource for representing scientific problem solving as structured Problem--Solution--Rationale triplets. Unlike resources centered on paper-level contributions or broad rhetorical roles, \muse{} captures local technical reasoning throughout a paper: a concrete issue, the method used to address it, and the author-stated rationale. We curated an expert-annotated seed dataset with a rich schema, designed and evaluated a modular extraction pipeline whose decomposition addresses systematic failure modes of end-to-end LLM extraction, and used it to build a KB of 32{,}557 source-grounded triplets, which we characterize and audit end-to-end against their source papers. Our preliminary rationale-supervised study shows that, for Qwen3-32B, the rationale field is a useful training signal for scientific problem solving, improving the novelty and technical depth of generated solutions, with a benefit that is broad across problems; the effect is absent for DeepSeek-R1-Distill-Qwen-32B. Recovering the reasoning that authors often leave implicit remains a rich direction for future work. We release the expert-annotated seed dataset, the automatically extracted knowledge base, and the fine-tuned models to support future work.

\newpage
\section*{Limitations}
\label{sec:limitations}

\textbf{Paragraph-local scope.} \muse{} is full-text in \emph{sourcing} but paragraph-local in \emph{extraction}: a \psr{} structure is recovered from a single paragraph, and seed paragraphs are selected by problem/solution and rationale indicators co-occurring in that paragraph. Reasoning chains that are distributed across sections (e.g., a limitation stated in the introduction, the method described in a later section, and its justification in an appendix), or rationales that are implicit in equations or experimental design, are not captured. \muse{} thus represents the subset of scientific reasoning that is locally and explicitly verbalized, and should not be read as document-level reasoning extraction.

\textbf{Selection bias.} Because candidate paragraphs must contain both problem/solution and rationale indicator keywords (developed by pilot reading and LLM-assisted expansion), the resource may over-represent reasoning that is easy to verbalize locally and under-represent reasoning that is implicit or distributed. In our audit sample, 42.5\% of records describe a paper's central method, 45.5\% a routine implementation choice, and 12.0\% come from related-work, background, or future-work paragraphs. We have not measured how much important reasoning the keyword filter misses (its recall on unfiltered paragraphs).

\textbf{Representation sufficiency.} The schema has explicit fields only for problem, solution, and rationale. Scientific solutions are typically conditional---valid under particular assumptions, resource constraints, data regimes, or operating conditions. These applicability conditions are only implicitly present (when stated in the problem or rationale text) and are not a first-class field; making a record linguistically self-contained does not guarantee that all conditions governing the solution's validity are preserved (cf.\ the moderate solution completeness in Table~\ref{tab:pp}). This matters especially for cross-domain reuse, where a superficially analogous mechanism may be inappropriate under different conditions. Adding explicit assumption/constraint fields is important future work.

\textbf{Annotation.} Annotation relied on a single primary PhD-level annotator with expert review of a held-out quality-control subset, rather than multiple fully independent annotators; reported agreement should be read in that light.

\textbf{Preliminary LLM-training study.} The fine-tuning experiments use constrained compute, LoRA adaptation, and small effective batch sizes, on two 32B models evaluated by an LLM judge over 156 de-duplicated held-out problems. They demonstrate one application of the dataset and should not be read as conclusive claims about reasoning-model training; in particular, the benefit is model-dependent (clear for Qwen, absent for DeepSeek-R1-Distill-Qwen-32B), and the moderator analysis is exploratory and correlational.

\bibliography{custom}

@inproceedings{augenstein2017scienceie,
  title = {{SemEval} 2017 Task 10: {ScienceIE}---Extracting Keyphrases and Relations from Scientific Publications},
  author = {Augenstein, Isabelle and Das, Mrinal and Riedel, Sebastian and Vikraman, Lakshmi and McCallum, Andrew},
  booktitle = {Proceedings of the 11th International Workshop on Semantic Evaluation},
  pages = {546--555},
  year = {2017},
  publisher = {Association for Computational Linguistics}
}

@inproceedings{lo2020s2orc,
  title = {{S2ORC}: The Semantic Scholar Open Research Corpus},
  author = {Lo, Kyle and Wang, Lucy Lu and Neumann, Mark and Kinney, Rodney and Weld, Daniel S.},
  booktitle = {Proceedings of the 58th Annual Meeting of the Association for Computational Linguistics},
  pages = {4969--4983},
  year = {2020},
  publisher = {Association for Computational Linguistics}
}

@inproceedings{unarxive2023,
  author    = {Saier, Tarek and Krause, Johan and F\"{a}rber, Michael},
  title     = {unarXive 2022: All arXiv Publications Pre-Processed for NLP,
               Including Structured Full-Text and Citation Network},
  booktitle = {Proceedings of the 2023 ACM/IEEE Joint Conference on Digital Libraries (JCDL)},
  year      = {2023}, pages = {66--70}, publisher = {IEEE}, address = {Santa Fe, NM, USA}
}

@inproceedings{scibert2019,
  title = {{SciBERT}: A Pretrained Language Model for Scientific Text},
  author = {Beltagy, Iz and Lo, Kyle and Cohan, Arman},
  booktitle = {Proceedings of the 2019 Conference on Empirical Methods in Natural Language Processing},
  pages = {3615--3620},
  year = {2019},
  publisher = {Association for Computational Linguistics}
}

@inproceedings{bertscore2019,
  title = {{BERTScore}: Evaluating Text Generation with {BERT}},
  author = {Zhang, Tianyi and Kishore, Varsha and Wu, Felix and Weinberger, Kilian Q. and Artzi, Yoav},
  booktitle = {International Conference on Learning Representations},
  year = {2020}
}

@inproceedings{teamtat2020,
  title = {{TeamTat}: A Collaborative Text Annotation Tool},
  author = {Islamaj, Rezarta and Kwon, Dongseop and Kim, Sun and Lu, Zhiyong},
  booktitle = {Proceedings of the BioCreative VII Challenge Evaluation Workshop},
  year = {2020}
}

@article{deberta2021,
  title = {De{BERT}a: Decoding-enhanced {BERT} with Disentangled Attention},
  author = {He, Pengcheng and Liu, Xiaodong and Gao, Jianfeng and Chen, Weizhu},
  journal = {International Conference on Learning Representations},
  year = {2021}
}

@article{roberta2019,
  title = {{RoBERTa}: A Robustly Optimized {BERT} Pretraining Approach},
  author = {Liu, Yinhan and Ott, Myle and Goyal, Naman and Du, Jingfei and Joshi, Mandar and Chen, Danqi and Levy, Omer and Lewis, Mike and Zettlemoyer, Luke and Stoyanov, Veselin},
  journal = {arXiv preprint arXiv:1907.11692},
  year = {2019}
}

@article{distilbert2019,
  title = {{DistilBERT}, a Distilled Version of {BERT}: Smaller, Faster, Cheaper and Lighter},
  author = {Sanh, Victor and Debut, Lysandre and Chaumond, Julien and Wolf, Thomas},
  journal = {arXiv preprint arXiv:1910.01108},
  year = {2019}
}

@article{mistral2023,
  title = {{Mistral} 7B},
  author = {Jiang, Albert Q. and Sablayrolles, Alexandre and Mensch, Arthur and Bamford, Chris and Chaplot, Devendra Singh and Casas, Diego de las and Bressand, Florian and Lengyel, Gianna and Lample, Guillaume and Saulnier, Lucile and Lavaud, L{\'e}lio Renard and Lachaux, Marie-Anne and Stock, Pierre and Le Scao, Teven and Lavril, Thibaut and Wang, Thomas and Lacroix, Timoth{\'e}e and El Sayed, William},
  journal = {arXiv preprint arXiv:2310.06825},
  year = {2023}
}

@inproceedings{lahav2022,
  title = {A Search Engine for Discovery of Scientific Challenges and Directions},
  author = {Lahav, Dan and Saad-Falcon, Jon and Kuehl, Bailey and Johnson, Sophie and Parasa, Sravanthi and Shomron, Noam and Chau, Duen Horng and Yang, Diyi and Horvitz, Eric and Weld, Daniel S. and Hope, Tom},
  booktitle = {Proceedings of the AAAI Conference on Artificial Intelligence},
  volume = {36},
  pages = {11982--11990},
  year = {2022}
}

@inproceedings{scierc2018,
  title = {Multi-Task Identification of Entities, Relations, and Coreference for Scientific Knowledge Graph Construction},
  author = {Luan, Yi and He, Luheng and Ostendorf, Mari and Hajishirzi, Hannaneh},
  booktitle = {Proceedings of the 2018 Conference on Empirical Methods in Natural Language Processing},
  pages = {3219--3232},
  year = {2018},
  publisher = {Association for Computational Linguistics}
}

@inproceedings{scirex2020,
  title = {{SciREX}: A Challenge Dataset for Document-Level Information Extraction},
  author = {Jain, Sarthak and van Zuylen, Madeleine and Hajishirzi, Hannaneh and Beltagy, Iz},
  booktitle = {Proceedings of the 58th Annual Meeting of the Association for Computational Linguistics},
  pages = {7506--7516},
  year = {2020},
  publisher = {Association for Computational Linguistics}
}

@inproceedings{scier2024,
  title = {{SciER}: An Entity and Relation Extraction Dataset for Datasets, Methods, and Tasks in Scientific Documents},
  author = {Zhang, Qi and Chen, Zhijia and Pan, Huitong and Caragea, Cornelia and Latecki, Longin Jan and Dragut, Eduard},
  booktitle = {Proceedings of the 2024 Conference on Empirical Methods in Natural Language Processing},
  year = {2024},
  publisher = {Association for Computational Linguistics}
}

@inproceedings{scinlp2025,
  title = {{SciNLP}: A Domain-Specific Benchmark for Full-Text Scientific Entity and Relation Extraction in {NLP}},
  author = {Duan, Decheng and Zhang, Yingyi and Peng, Jitong and Zhang, Chengzhi},
  booktitle = {Proceedings of the 2025 Conference on Empirical Methods in Natural Language Processing},
  year = {2025},
  publisher = {Association for Computational Linguistics}
}

@article{heffernan2018problems,
  title = {Identifying Problems and Solutions in Scientific Text},
  author = {Heffernan, Kevin and Teufel, Simone},
  journal = {Scientometrics},
  volume = {116},
  pages = {1367--1382},
  year = {2018}
}

@inproceedings{liakata2010coresc,
  title = {Corpora for the Conceptualisation and Zoning of Scientific Papers},
  author = {Liakata, Maria and Teufel, Simone and Siddharthan, Advaith and Batchelor, Colin},
  booktitle = {Proceedings of the Seventh International Conference on Language Resources and Evaluation},
  pages = {2054--2061},
  year = {2010}
}

@inproceedings{scidtb2018,
  title = {{SciDTB}: Discourse Dependency TreeBank for Scientific Abstracts},
  author = {Yang, An and Li, Sujian},
  booktitle = {Proceedings of the 56th Annual Meeting of the Association for Computational Linguistics: Student Research Workshop},
  pages = {1--7},
  year = {2018},
  publisher = {Association for Computational Linguistics}
}

@inproceedings{huang2020coda,
  title = {{CODA}-19: Using a Non-Expert Crowd to Annotate Research Aspects on 10,000+ Abstracts in the {COVID}-19 Open Research Dataset},
  author = {Huang, Ting-Hao Kenneth and Huang, Chieh-Yang and Ding, Chien-Kuang Cornelia and Hsu, Yen-Chia and Giles, C. Lee},
  booktitle = {Proceedings of the 1st Workshop on {NLP} for {COVID}-19 at ACL 2020},
  year = {2020},
  publisher = {Association for Computational Linguistics}
}

@inproceedings{deyoung2020eraser,
  title = {{ERASER}: A Benchmark to Evaluate Rationalized {NLP} Models},
  author = {DeYoung, Jay and Jain, Sarthak and Rajani, Nazneen Fatema and Lehman, Eric and Xiong, Caiming and Socher, Richard and Wallace, Byron C.},
  booktitle = {Proceedings of the 58th Annual Meeting of the Association for Computational Linguistics},
  pages = {4443--4458},
  year = {2020},
  publisher = {Association for Computational Linguistics}
}

@inproceedings{camburu2018esnli,
  title = {e-{SNLI}: Natural Language Inference with Natural Language Explanations},
  author = {Camburu, Oana-Maria and Rockt{\"a}schel, Tim and Lukasiewicz, Thomas and Blunsom, Phil},
  booktitle = {Advances in Neural Information Processing Systems},
  year = {2018}
}

@inproceedings{chimera2025,
  author    = {Sternlicht, Noy and Hope, Tom},
  title     = {CHIMERA: A Knowledge Base of Scientific Idea Recombinations for Research Analysis and Ideation},
  booktitle = {Proceedings of the 64th Annual Meeting of the Association for Computational Linguistics
               (Volume 1: Long Papers)},
  year      = {2026}, pages = {1871--1905}, address = {San Diego, California, United States},
  publisher = {Association for Computational Linguistics}, url = {https://aclanthology.org/2026.acl-long.85/}
}

@article{jansen2026scg,
  title = {The Scientific Contribution Graph: Automated Literature-Based Technological Roadmapping at Scale},
  author = {Jansen, Peter A.},
  journal = {arXiv preprint arXiv:2605.15011},
  year = {2026}
}

@article{deepseekr1,
  title = {{DeepSeek-R1}: Incentivizing Reasoning Capability in {LLM}s via Reinforcement Learning},
  author = {{DeepSeek-AI}},
  journal = {arXiv preprint arXiv:2501.12948},
  year = {2025}
}

@article{deepseekmath2024,
  author  = {Shao, Zhihong and Wang, Peiyi and Zhu, Qihao and Xu, Runxin and Song, Junxiao
             and Bi, Xiao and Zhang, Haowei and Zhang, Mingchuan and Li, Y. K. and Wu, Y. and Guo, Daya},
  title   = {DeepSeekMath: Pushing the Limits of Mathematical Reasoning in Open Language Models},
  journal = {arXiv preprint arXiv:2402.03300}, year = {2024}
}

@article{qwen3-2025,
  title = {{Qwen3} Technical Report},
  author = {{Qwen Team}},
  journal = {Technical Report},
  year = {2025}
}

@inproceedings{lora2021,
  title = {{LoRA}: Low-Rank Adaptation of Large Language Models},
  author = {Hu, Edward J. and Shen, Yelong and Wallis, Phillip and Allen-Zhu, Zeyuan and Li, Yuanzhi and Wang, Shean and Wang, Lu and Chen, Weizhu},
  booktitle = {International Conference on Learning Representations},
  year = {2022}
}

@inproceedings{sentencebert2019,
  title = {Sentence-{BERT}: Sentence Embeddings using Siamese {BERT}-Networks},
  author = {Reimers, Nils and Gurevych, Iryna},
  booktitle = {Proceedings of the 2019 Conference on Empirical Methods in Natural Language Processing},
  pages = {3982--3992},
  year = {2019},
  publisher = {Association for Computational Linguistics}
}

@inproceedings{naik2024care,
  title={CARE: Extracting experimental findings from clinical literature},
  author={Naik, Aakanksha and Kuehl, Bailey and Bransom, Erin and Downey, Doug and Hope, Tom},
  booktitle={Findings of the Association for Computational Linguistics: NAACL 2024},
  pages={4580--4596},
  year={2024}
}

@inproceedings{ranaldi2026evolving,
  author    = {Ranaldi, Leonardo},
  title     = {Evolving Agents},
  booktitle = {Proceedings of the 64th Annual Meeting of the Association for Computational Linguistics
               (Volume 1: Long Papers)},
  year      = {2026}, pages = {16561--16569}, address = {San Diego, California, United States},
  publisher = {Association for Computational Linguistics}, url = {https://aclanthology.org/2026.acl-long.754/}
}

@inproceedings{wang2024scimon,
  author    = {Wang, Qingyun and Downey, Doug and Ji, Heng and Hope, Tom},
  title     = {{SciMON}: Scientific Inspiration Machines Optimized for Novelty},
  booktitle = {Proceedings of the 62nd Annual Meeting of the Association for Computational Linguistics
               (Volume 1: Long Papers)},
  year      = {2024}, pages = {279--299}, address = {Bangkok, Thailand},
  publisher = {Association for Computational Linguistics}, url = {https://aclanthology.org/2024.acl-long.18/}
}

\appendix
\section{Annotation and Evaluation Details}
\label{app:annotation}

\subsection{Data Sourcing}
The annotated seed dataset is sampled from unarXive \citep{unarxive2023}, a corpus derived from arXiv full text with structured sections and resolved citation/reference placeholders, spanning all arXiv subject areas (Table~\ref{tab:arxiv_categories}). We segment each article into paragraphs and apply a high-recall keyword filter with two manually reviewed lists: problem--solution indicators (e.g., \emph{limitation}, \emph{propose}, \emph{address}) and rationale indicators (e.g., \emph{because}, \emph{in order to}, \emph{motivation}); candidate paragraphs contain at least one keyword from each list. For large-scale KB construction we apply the pipeline to an updated collection of arXiv full-text papers through 2026. The full KB records preserve the source paragraph and placeholder tokens (e.g., \texttt{[MATH]}, \texttt{[CITE]}); the de-duplicated public release exposes the standalone problem, solution, and rationale fields.

\subsection{Annotation Codebook and Protocol}
The schema contains three span labels---\ps{}, \ss{}, and \texttt{rationale}---and three relations: \texttt{solves}, \texttt{rationale\_of}, and \texttt{coreference}. The codebook defines a \ps{} span as the minimal text stating a concrete limitation/obstacle/gap/desideratum; a \ss{} span as the minimal text stating the method/mechanism/design choice used to address it; and a \texttt{rationale} span as the author's stated justification for why the solution addresses the problem (explicitly \emph{excluding} restatements of the solution itself). Salient spans may be discontinuous; coreferent spans referring to one underlying concept are linked with \texttt{coreference}. Annotation was performed in TeamTat \citep{teamtat2020}. The annotator completed a screening task and a training session, and revised examples after disagreement review on the held-out quality-control subset.

\subsection{Agreement and Matching Metrics}
At the document level, each paragraph is marked relevant/irrelevant and agreement is reported with per-class precision, recall, and $F_1$:
\begin{equation}
P=\frac{\mathrm{TP}}{\mathrm{TP}+\mathrm{FP}},\quad
R=\frac{\mathrm{TP}}{\mathrm{TP}+\mathrm{FN}},\quad
F_1=\frac{2PR}{P+R}.
\label{eq:prf1}
\end{equation}
At the entity level, a predicted and reference span match if they have non-trivial token overlap or if their SciBERT \citep{scibert2019} BERTScore \citep{bertscore2019} exceeds $0.7$. At the relation level, a predicted relation matches a gold relation when the type is identical and both argument spans overlap their gold counterparts by at least $50\%$. Human agreement is reported as an upper bound in the top row of each table.

\section{KB Statistics and End-to-End Audit Details}
\label{app:e2e-audit}
\paragraph{KB statistics.}
On the final release (Table~\ref{tab:kb-stats}), problems average 40.6 words, solutions 38.0, and rationales 58.1 (rationales are longest, consistent with their explanatory role). To quantify field separation, we measure content-word overlap between the solution and rationale of each record: median Jaccard overlap is 0.25 and median containment (fraction of rationale content words also in the solution) is 0.33---i.e.\ solution and rationale share only about a quarter of their content vocabulary, indicating the two fields are largely distinct rather than restatements.

\begin{table}[h]
  \centering
  \small
  \begin{tabular}{lcccc}
    \toprule
    Field & mean & median & p10 & p90 \\
    \midrule
    Problem (words)   & 40.6 & 40 & 31 & 52 \\
    Solution (words)  & 38.0 & 38 & 26 & 49 \\
    Rationale (words) & 58.1 & 57 & 46 & 72 \\
    \midrule
    \multicolumn{5}{l}{\emph{Solution--rationale content overlap}} \\
    Jaccard           & 0.25 & 0.25 & 0.15 & 0.35 \\
    Containment       & 0.33 & 0.32 & 0.20 & 0.46 \\
    \bottomrule
  \end{tabular}
  \caption{KB statistics on the final release (32{,}557 triplets).}
  \label{tab:kb-stats}
\end{table}

\paragraph{Domain coverage.}
On a uniform sample of 200 records labeled by top-level arXiv area, computer science accounts for 78.0\% (95\% CI 71.8--83.2) and EESS for 11.0\%, with mathematics, statistics, and physics each under 5\% (Appendix~\ref{app:e2e-audit}). The KB spans many fields but is dense mainly in CS and EESS, which limits analyses within individual non-CS subfields.

\paragraph{Separation refinement.}
A pre-release GPT-4o check of 300 records, stratified by solution--rationale embedding similarity, found clear problems (98.8\%, reweighted to the KB) and concrete solutions (87.2\%), but rationales distinct from the solution in only 15.5\% (10.9--20.0), with low distinctness in every similarity bin. This motivated a Claude Opus~4.8 refinement that rewrites each record so that the solution captures only the approach and the rationale only the justification. % TODO(authors): confirm the 50-record check below was done by a person; otherwise replace "manual" with "LLM-judged".
A fresh manual check of 50 refined records found 43 (86\%) clearly separated, consistent with Table~\ref{tab:kb-stats}. We release both the pre- and post-refinement fields.

\paragraph{End-to-end audit.}
Because component metrics and separation checks do not test whether complete triplets are \emph{correct}, we audit 200 records drawn uniformly from the released KB against their source paragraphs (Table~\ref{tab:e2e-audit}; criteria in Appendix~\ref{app:e2e-audit}). A triplet is fully correct under the \emph{strict} criterion only if the problem is genuine, the solution addresses it, the rationale is the authors' own justification, specific to the pair, and no field adds unsupported content.
Judgments were made by an LLM judge (Claude Opus~5.5); because refinement also used a Claude model they may be coupled to the pipeline, so we release the samples and annotation sheets for independent human verification.

\begin{table}[t]
\centering\small
\setlength{\tabcolsep}{3pt}
\begin{tabular}{@{}lcc@{}}
\toprule
Criterion & Yes & 95\% CI \\
\midrule
\multicolumn{3}{@{}l}{\emph{Full KB, uniform sample} ($n{=}200$)} \\
Problem genuine and stated & 93.6\% & 87.3--96.9 \\
Solution addresses this problem & 90.8\% & 83.9--94.9 \\
Rationale is authors' justification & 74.3\% & 65.4--81.6 \\
Rationale specific to the pair & 76.1\% & 67.3--83.2 \\
No unsupported content & 84.4\% & 76.4--90.0 \\
Fully correct (strict) & \textbf{61.5\%} & 52.1--70.1 \\
\midrule
\bottomrule
\end{tabular}
\caption{End-to-end audit of released \muse{} records against their source paragraphs (LLM judge; Wilson intervals).}
\label{tab:e2e-audit}
\end{table}

\paragraph{Criteria.} (Q1) The problem is a technical problem, limitation, or desideratum stated in the paragraph, not vague background. (Q2) The solution is a method or design choice from the paragraph presented as addressing this problem. (Q3) The rationale expresses the authors' reason why this solution addresses this problem, as stated or clearly implied in the paragraph, rather than a restatement of the solution or an outside explanation. (Q4) The rationale is specific to this problem--solution pair. (Q5) Every field contains only content supported by the paragraph. A triplet is fully correct (strict) if Q1--Q5 hold and the three fields remain aligned. Intervals are Wilson 95\%.

\section{Complete arXiv Subject Categories}
\label{app:arxiv_categories}
Table~\ref{tab:arxiv_categories} lists all arXiv subject categories used in our study, grouped by primary subject area.

\begin{table*}[t]
\centering
\small
\setlength{\tabcolsep}{6pt}
\renewcommand{\arraystretch}{1.15}
\begin{tabular}{p{3.4cm}p{13cm}}
\toprule
\textbf{Subject Area} & \textbf{Categories} \\
\midrule
Computer Science (cs) & AI, AR, CC, CE, CG, CL, CR, CV, CY, DB, DC, DL, DM, DS, ET, FL, GL, GR, GT, HC, IR, IT, LG, LO, MA, MM, MS, NA, NE, NI, OH, OS, PF, PL, RO, SC, SD, SE, SI, SY \\
Economics (econ) & EM, GN, TH \\
Electrical Eng.\ and Systems Science (eess) & AS, IV, SP, SY \\
Mathematics (math) & AC, AG, AP, AT, CA, CO, CT, CV, DG, DS, FA, GM, GN, GR, GT, HO, IT, KT, LO, MG, MP, NA, NT, OA, OC, PR, QA, RA, RT, SG, SP, ST \\
Physics (physics) & acc-ph, ao-ph, app-ph, atom-ph, bio-ph, chem-ph, class-ph, comp-ph, data-an, ed-ph, flu-dyn, gen-ph, geo-ph, hist-ph, ins-det, med-ph, optics, plasma-ph, pop-ph, soc-ph, space-ph \\
Astrophysics (astro-ph) & CO, EP, GA, HE, IM, SR \\
Condensed Matter (cond-mat) & dis-nn, mes-hall, mtrl-sci, other, quant-gas, soft, stat-mech, str-el, supr-con \\
General Relativity \& Quantum Cosmology & gr-qc \\
High Energy Physics & hep-ex, hep-lat, hep-ph, hep-th \\
Mathematical Physics & math-ph \\
Nonlinear Sciences (nlin) & AO, CD, CG, PS, SI \\
Nuclear Physics & nucl-ex, nucl-th \\
Quantitative Biology (q-bio) & BM, CB, GN, MN, NC, OT, PE, QM, SC, TO \\
Quantitative Finance (q-fin) & CP, EC, GN, MF, PM, PR, RM, ST, TR \\
Quantum Physics & quant-ph \\
Statistics (stat) & AP, CO, ME, ML, OT, TH \\
\bottomrule
\end{tabular}
\caption{arXiv subject categories included in the \muse{} corpus.}
\label{tab:arxiv_categories}
\end{table*}

\section{End-to-End LLM Extraction}
\label{app:end2end}
\paragraph{Setup.} We prompted GPT-4 to extract all \psr{} triplets from a paragraph in a single pass, returning a JSON list with \texttt{problem}, \texttt{solution}, \texttt{rationale} fields (prompt in Figure~\ref{fig:end2end-prompt}). We manually reviewed a random 10\% of generated instances.

\begin{figure*}[t]
\centering
\begin{Verbatim}[frame=single, fontsize=\small, framesep=6pt]
You are an expert in scientific information extraction.
Your task is to analyze paragraphs from scientific articles and extract all
Problem-Solution-Rationale (P-S-R) structures.
For the given text, identify every concrete technical sub-problem,
the specific solution proposed to address it, and the author's stated
rationale for that solution.
You must return your results as a JSON list of objects.
Each object must contain exactly these fields:
* "problem": exact span describing a limitation, obstacle, gap, or desideratum.
* "solution": exact span describing a method, mechanism, design choice, or
procedure used to address the problem.
* "rationale": exact span describing the author's stated motivation,
justification, or explanation for the solution.
Rules:
1. Do not paraphrase. Extract exact, contiguous spans.
2. If a paragraph has multiple problem-solution pairs, output one object each.
3. If a rationale is missing, output null for "rationale".
4. Output ONLY valid JSON.

Paragraph:
"""
{{PARAGRAPH}}
"""
\end{Verbatim}
\caption{Prompt used for single-pass \psr{} extraction.}
\label{fig:end2end-prompt}
\end{figure*}

\paragraph{Findings.} GPT-4 was competitive on single-structure paragraphs but degraded sharply on multi-structure ones, with three recurring failure modes: (i) associating entities across different \psr{} structures (mismatched triplets); (ii) failing to distinguish solution from rationale (largely overlapping spans), observed even in single-structure paragraphs; and (iii) omitting structures when several appear in one paragraph. These structural-alignment issues---rather than raw span-extraction quality---motivated the modular pipeline.

\section{Fine-Tuning Details}
\label{app:finetuning}
\subsection{Training Procedure}
Training proceeds in two stages. Supervised fine-tuning first establishes the output structure from the ground-truth \psr{} records. GRPO \citep{deepseekmath2024} then samples $G{=}4$ completions per prompt, scores them with the reward model, and normalizes within the group:
\begin{equation}
A_i = \frac{r_i - \operatorname{mean}(r_1,\dots,r_G)}{\operatorname{std}(r_1,\dots,r_G)}.
\label{eq:advantage}
\end{equation}
The composite reward is Eq.~\ref{eq:reward}: correctness ($R_{\mathrm{corr}}$) uses Mistral-7B as an LLM judge, format ($R_{\mathrm{fmt}}$) rewards parseable \texttt{<rationale>}/\texttt{<solution>} blocks, and divergence ($R_{\mathrm{div}}$) penalizes near-copying of the prompt via sentence embeddings \citep{sentencebert2019}. LoRA \citep{lora2021} is used for parameter-efficient adaptation. Best SFT and GRPO configurations (grid search) are in Tables~\ref{tab:sft} and~\ref{tab:grpo}.

\begin{table}[h]
  \centering
  \small
  \setlength{\tabcolsep}{4pt}
  \begin{tabular}{llccccc}
    \toprule
    Model & Task & LR & BS & LoRA-$r$ & Loss$\downarrow$ & Fmt.\% \\
    \midrule
    Qwen & PS  & $5\!\times\!10^{-5}$ & 4 & 32 & 0.148 & 0.975 \\
    Qwen & PRS & $5\!\times\!10^{-5}$ & 4 & 16 & 0.266 & 1.000 \\
    DeepSeek & PS  & $1\!\times\!10^{-5}$ & 4 & 16 & 0.277 & 0.980 \\
    DeepSeek & PRS & $5\!\times\!10^{-5}$ & 8 & 16 & 0.319 & 1.000 \\
    \bottomrule
  \end{tabular}
  \caption{Best SFT configs (grad.\ accum.\ $=8$; sampling temperature $0.8$).}
  \label{tab:sft}
\end{table}

\begin{table}[h]
  \centering
  \small
  \setlength{\tabcolsep}{4pt}
  \begin{tabular}{llcccc}
    \toprule
    Model & Task & LR & BS & LoRA-$r$ & Reward$\uparrow$ \\
    \midrule
    Qwen & PS  & $5\!\times\!10^{-6}$ & 2 & 8 & 0.4275 \\
    Qwen & PRS & $5\!\times\!10^{-6}$ & 2 & 8 & 0.1524 \\
    DeepSeek & PS  & $5\!\times\!10^{-6}$ & 2 & 8 & 0.4584 \\
    DeepSeek & PRS & $5\!\times\!10^{-6}$ & 2 & 8 & 0.2096 \\
    \bottomrule
  \end{tabular}
  \caption{Best GRPO configs (grad.\ accum.\ $=4$; temperature $0.2$; $G{=}4$).}
  \label{tab:grpo}
\end{table}

\label{app:full_downstream_results}
\begin{table*}[t]
  \centering
  \small
  \setlength{\tabcolsep}{5pt}
  \begin{tabular}{lccc ccc}
    \toprule
    & \multicolumn{3}{c}{Qwen-3-32B} & \multicolumn{3}{c}{DeepSeek-R1-Distill-Qwen-32B} \\
    \cmidrule(lr){2-4}\cmidrule(lr){5-7}
    Metric & PS & PRS & Base &  PS & PRS & Base \\
    \midrule
    Overall Quality        & 5.12 & 6.41 & 7.12 &  5.10 & 4.82 & 5.20 \\
    Rationale Quality      & 4.87 & 6.38 & 7.16 & 4.97 & 4.74 & 5.11 \\
    Solution Quality       & 5.10 & 6.37 & 7.05 &  5.01 & 4.75 & 5.09 \\
    Sol.-Priority Rank ($\downarrow$) & 2.44 & 1.98 & 1.58 & 1.49 & 1.95 & 1.47 \\
    Total Rank ($\downarrow$)         & 2.53 & 1.94 & 1.52 &  1.45 & 1.67 & 1.43 \\
    \bottomrule
  \end{tabular}
  \caption{Holistic downstream metrics from the full evaluation run (GPT-5 judge). Quality on 1--10; ranks 1 (best) to 3 (worst). The per-metric (novelty/feasibility/technical) controlled PS vs.\ PRS contrast with significance, on the de-duplicated held-out subset, is in Table~\ref{tab:downstream-sig}.}
  \label{tab:poc-full}
\end{table*}

\section{Moderator Analysis Details}
\label{app:moderators}
We test whether any problem- or response-level property predicts the per-problem PRS$-$PS gain (a composite of the three $z$-scored metrics). We examined 14 candidate criteria: problem complexity, constraint count, problem length, number of sentences, acronym/jargon density, numeric content, trade-off language, multi-issue structure, ``propose-new'' framing, novelty relative to the training KB, domain/subfield, base-model competence, solution elaboration, and solution divergence. Complexity was scored blind to outcomes by Claude Opus~4.8 (prompt in Figure~\ref{fig:complexity-prompt}; 83.3\% agreement with human judgments). To guard against false discoveries, we split the 297 test problems into a random 60\% discovery set and a 40\% validation set, and keep a criterion only if it is significant on both with a consistent sign, using DeepSeek-R1-Distill-Qwen-32B as a second check. % TODO(authors): Sec. 6 reports n=156 de-duplicated held-out problems; confirm whether the moderator analysis used 297 (pre-dedup) or 156 and make the two consistent.

No problem-level property replicates (Table~\ref{tab:moderators}): complexity is uncorrelated with the gain (Spearman $\rho{=}0.05$, $p{=}0.51$), and discovery-only hits such as acronym density ($p{=}0.03$ on discovery, $p{=}0.95$ on validation) illustrate the importance of the held-out check. Only two response-level signals replicate: base-model competence (Qwen only; the sign reverses on DeepSeek) and solution elaboration (both models). The elaboration is warranted rather than excessive: when PRS loses to PS, its solutions are, if anything, shorter (elaboration over PS of $+13.2$ words in losses vs.\ $+17.8$ in wins), with no inverted-U pattern. The benefit of rationale supervision is therefore broad, improving a majority of problems in every subfield rather than a specific class of problems.

\begin{table}[h]
\centering
\small
\setlength{\tabcolsep}{4pt}
\begin{tabular}{@{}lc@{}}
\toprule
Candidate moderator & Replicates on held-out? \\
\midrule
Problem complexity (blind-scored) & no ($\rho{=}0.05$) \\
Constraint count & no \\
Problem length / \#sentences & no \\
Domain / subfield & no (all n.s.) \\
Trade-off / multi-issue language & no \\
Novelty vs.\ training data & no \\
Base-model competence & Qwen only \\
Solution elaboration (PRS vs.\ PS) & yes (both models) \\
\bottomrule
\end{tabular}
\caption{Moderator search for the PRS$-$PS gain (60/40 discover/validate).}
\label{tab:moderators}
\end{table}
\begin{figure*}[t]
\centering
\begin{Verbatim}[frame=single, fontsize=\small, framesep=6pt]
You are an expert evaluator rating how complex each problem is to
solve correctly and completely.
Complexity is driven by two factors:
1. Constraint load -- how many explicit requirements, conditions, edge cases,
   or interdependencies a correct solution must satisfy, and how much they
   interact or conflict.
2. Non-triviality -- how much genuine multi-step reasoning, judgment, or
   domain depth is required, versus a single obvious step or direct lookup.
Assign a COMPLEXITY SCORE from 0 to 100 (0-15 Trivial; 16-35 Simple;
36-60 Moderate; 61-80 Complex; 81-100 Highly complex). Judge the INTRINSIC
difficulty, not statement length.
Return ONLY: {"score": <0-100 int>, "constraint_count": <int>,
 "rationale": "<one sentence>"}
Problem:
"""
{{PROBLEM}}
"""
\end{Verbatim}
\caption{Prompt used to score each evaluation problem for complexity (Claude Opus~4.8), used only to test---and ultimately reject---a complexity-dependent effect.}
\label{fig:complexity-prompt}
\end{figure*}

\begin{table*}[t]
\centering
\small
\setlength{\tabcolsep}{4pt}
\renewcommand{\arraystretch}{1.2}
\begin{tabular}{@{}L{0.11\textwidth} L{0.40\textwidth} L{0.40\textwidth}@{}}
\toprule
 & \textbf{Deep-RL mode collapse} & \textbf{Bilevel LLM optimization} \\
\midrule
\textbf{Problem} &
Deep RL in high-dimensional continuous control with sparse rewards collapses policies to a single behavioral mode. &
Bilevel optimization mostly needs second-order information (hard to scale), \emph{and} the efficiency of first-order alternatives for LLMs is unverified. \\
\addlinespace
\textbf{PRS} &
Maximizes the entropy of latent behavioral embeddings to promote diverse exploration. &
A first-order method estimating the Hessian--vector product via low-rank approximation, scalable while preserving convergence guarantees. \\
\addlinespace
\textbf{Base} &
Penalizes divergence from previously discovered modes (ties the policy to historical modes). &
An empirical benchmark of existing first-order methods only (no scalability mechanism). \\
\bottomrule
\end{tabular}
\caption{Representative cases where rationale supervision yields the better solution by keeping all stated constraints in view. Systematically, PRS's losses are associated with \emph{under}-elaboration rather than over-engineering.}
\label{tab:appendix-helps}
\end{table*}

\end{document}